\documentclass[conference, a4paper]{IEEEtran}

\usepackage{ifpdf}

\usepackage{cite}

\ifCLASSINFOpdf
 \usepackage[pdftex]{graphicx}
\else
\fi

\usepackage{algorithmic}

\usepackage{array}

\usepackage{url}

\usepackage{graphicx}
\usepackage{fixltx2e}
\usepackage{graphicx}
\usepackage{tabularx}
\usepackage{soul}
\usepackage{epstopdf}
\usepackage[]{inputenc}
\usepackage{hyperref}
\usepackage{xstring}
\usepackage{mathptmx}
\usepackage{graphics}
\usepackage{multirow}
\usepackage{amsmath}
\usepackage{tabularx}
\usepackage{diagbox}
\usepackage{multicol}
\usepackage{enumitem}
\usepackage{listings}
\usepackage{bm}
\usepackage{textcomp}
\newcommand{\RNum}[1]{\uppercase\expandafter{\romannumeral #1\relax}}

\usepackage{enumitem}

\newcounter{SentenceCounter}
\newcounter{ExampleCounter}

\newenvironment{Sentence}[1][]{
\refstepcounter{SentenceCounter}
\vspace{0.5mm}
\def\temp{#1}\ifx\temp\empty
    \noindent Sentence~\theExampleCounter.\theSentenceCounter:
\else
    \noindent Sentence~\theExampleCounter.\theSentenceCounter\hspace{1mm}(#1):
\fi
\itshape
}
{
\label{sent\theExampleCounter.\theSentenceCounter}
}

\newenvironment{SentenceExample}[1][]{
\setcounter{SentenceCounter}{0}
\vspace{0.5mm}
\begin{table}[htb]
\centering
\bgroup
\def\arraystretch{0.4}
\begin{tabular}{|p{0.45\textwidth}|}
\hline
\scriptsize
\vspace{1mm}
\def\temp{#1}
\ifx\temp\empty
    \refstepcounter{ExampleCounter}
    \noindent\textbf{Example \theExampleCounter}
\else
    \noindent\textbf{Example \theExampleCounter\hspace{1mm}(#1)}
\fi
\footnotesize
\begin{itemize}[leftmargin=*]

}{
\end{itemize}
\\\hline
\end{tabular}
\egroup
\label{eg\theExampleCounter}
\end{table}
\vspace{0.5mm}
}

\begin{document}
%

\title{SigmaLaw-ABSA: Dataset for Aspect-Based Sentiment Analysis in Legal Opinion Texts}

\author{\IEEEauthorblockN{
Chanika Ruchini Mudalige\IEEEauthorrefmark{1}\IEEEauthorrefmark{1},
Dilini Karunarathna\IEEEauthorrefmark{1}\IEEEauthorrefmark{2},
Isanka Rajapaksha\IEEEauthorrefmark{1}\IEEEauthorrefmark{3},\\ 
Nisansa de Silva\IEEEauthorrefmark{1}\IEEEauthorrefmark{4}, 
Gathika Ratnayaka\IEEEauthorrefmark{1}\IEEEauthorrefmark{5}, 
Amal Shehan Perera\IEEEauthorrefmark{1}\IEEEauthorrefmark{6} and
Ramesh Pathirana\IEEEauthorrefmark{2}\IEEEauthorrefmark{7}}

\IEEEauthorblockA{\IEEEauthorrefmark{1}Department of Computer Science \& Engineering,
University of Moratuwa}
\IEEEauthorblockA{Email: chanikaruchini.16@cse.mrt.ac.lk\IEEEauthorrefmark{1}, dilinirasanjana.16@cse.mrt.ac.lk\IEEEauthorrefmark{2},israjapaksha.16@cse.mrt.ac.lk\IEEEauthorrefmark{3},\\
NisansaDdS@cse.mrt.ac.lk\IEEEauthorrefmark{4},
gathika.14@cse.mrt.ac.lk\IEEEauthorrefmark{5},
shehan@cse.mrt.ac.lk\IEEEauthorrefmark{6}}
\IEEEauthorblockA{\IEEEauthorrefmark{2} Faculty of Law, University of Colombo}
\IEEEauthorblockA{Email: hp95ramesh@gmail.com\IEEEauthorrefmark{7}}
}


\maketitle

\begin{abstract}
Aspect-Based Sentiment Analysis (ABSA) has been prominent and ongoing research over many different domains, but it is not widely discussed in the legal domain. A number of publicly available datasets for a wide range of domains usually fulfill the needs of researchers to perform their studies in the field of ABSA. To the best of our knowledge, there is no publicly available dataset for the Aspect (Party) Based Sentiment Analysis for legal opinion texts. Therefore, creating a publicly available dataset for the research of ABSA for the legal domain can be considered as a task with significant importance. In this study, we introduce a manually annotated legal opinion text dataset (SigmaLaw-ABSA) intended towards facilitating researchers for ABSA tasks in the legal domain. SigmaLaw-ABSA consists of legal opinion texts in the English language which have been annotated by human judges. This study discusses the sub-tasks of ABSA relevant to the legal domain and how to use the dataset to perform them. This paper also describes the statistics of the dataset and as a baseline, we present some results on the performance of some existing deep learning based systems on the SigmaLaw-ABSA dataset.
\end{abstract}

 \renewcommand\IEEEkeywordsname{Keywords}
 \begin{IEEEkeywords}
 Aspect-Based Sentiment Analysis, Legal Domain, Information extraction, Natural Language Processing 
 \end{IEEEkeywords}

%
\IEEEpeerreviewmaketitle

\section{Introduction}
Legal opinion texts bear significant importance to legal officials as they contain valuable legal information such as arguments, legal opinions, and judgments. Such information is extremely valuable, specially when it comes to handling new legal cases \cite{inbook,Jayawardana2017DerivingAR}. However, the process of manually obtaining such information from legal opinion texts demands significant effort and time. Therefore, it is important to automate information extraction from legal opinion texts when it is needed to transform the legal domain using Artificial Intelligence \cite{inbook}. Extracting the views or opinions expressed in the legal texts is an important task of the process of automating information extraction within the legal domain. In the field of Natural Language Processing (NLP), Sentiment Analysis plays a major role when identifying the opinion in texts. A legal case usually consists of two or more parties and in the legal domain, an individual who only appears as a witness in the court case is not considered as a party. A party is an individual, an organization, a group of individuals or a group of both that composes a single entity. Thus a single text typically has several legal parties that express themselves independently. Extracting separate opinions of each legal party mentioned in a sentence could not be achieved using only the sentiment analysis process. This is where Aspect-Based Sentiment Analysis comes to play in the legal domain. 

Aspect-Based Sentiment Analysis (ABSA) is a text analysis technique that separates text into aspects, and then allocates each one a sentiment level (\textit{positive}, \textit{negative}, or \textit{neutral}) \cite{Bhoi2018VariousAT}. ABSA has been prominent and ongoing research in various domains. However, sentiment analysis and Aspect-Based Sentiment Analysis are not widely discussed in the legal domain. Though that is the case, they carry significant importance when referring to case law. In the legal domain, ABSA techniques can be used  to identify the level of sentiment in a sentence relevant to each legal party considering the legal parties as aspects. When a single sentence in a legal document has multiple legal parties, the sentiment of the sentence should be addressed relevant to each party. In the legal domain, sentiment analysis has been carried out only at a phrase/sentence level\cite{gamage2018fast}. Sentiment analysis  may not be sufficient to figure out the sentiment level of sentences for a particular party.

\begin{SentenceExample}
\item \begin{Sentence}In 2008, federal officials received a tip from a confidential informant that Lee had sold the informant ecstasy and marijuana.\end{Sentence}
\label{ex1}
\end{SentenceExample}

Consider Example~\ref{ex1} taken from Lee v. United States~\cite{1977lee} which consists of two legal parties; petitioner, Lee and defendant, government. Here, federal officials represent the government. The sentence in the example mentioned that officials received a tip about illegal actions of Lee. When considering the context of this sentence, we can clearly see that the context has positive sentiment regarding the government and negative sentiment regarding the person Lee.

There are many studies for the Aspect-Based Sentiment Analysis in other domains except the legal domain. Hotels, restaurants, movies, and product reviews are some publicly accessible datasets which typically satisfy the requirements of the researchers to perform their studies in the field of ABSA~\cite{alqaryouti2018sentiment}. However, it is not much effective to use a dataset created for a particular domain to evaluate ABSA tasks in a different domain. This implies that domain-specific datasets need to be created to evaluate ABSA tasks. The objective of our study is to create a publicly available dataset for research of the Party-Based Sentiment Analysis for legal opinion texts. This paper introduces a human-annotated dataset prepared to support various research tasks, including aspect extraction, polarity detection, party identification, and party- polarity detection. 

The structure of the paper is organized as follows. The related work in the literature is described in the Section \RNum{2}. Section \RNum{3} describes the process of data collection and annotation. Section \RNum{4} describes the research directions of the ABSA field relevant to the legal domain. Section \RNum{5} illustrates the statistics of the SigmaLaw-ABSA dataset. Section \RNum{6} describes the experiments that we followed with some existing deep learning based systems and finally, section \RNum{7} concludes the discussion with our future plans and works.

\section{Related Work}

\subsection{Legal Information Extraction}

Legal Domain is a domain which can be made more productive and effective with the introduction of Artificial Intelligence. This fact is evidenced by the number of studies conducted over many years on automatic legal information extraction. There have recently been experimented on calculating similarity measures in legal texts\cite{sugathadasa2017synergistic} by using \textit{Large Legal Text Corpus and Word Embeddings dataset} which consists of Legal Case Corpus, word2vec models of the legal domain, and several gazetteer lists. For the task of building legal information retrieval systems Sugathadasa et al. \cite{inbook} created \textit{Legal Information Retrieval dataset} using \textit{Large Legal Text Corpus and Word Embeddings dataset}. Further, it consists of created legal ontologies, gazetteer lists of the legal domain and result vectors obtained by this study. Moreover, few studies on the construction of legal ontology have been published\cite{Jayawardana2017DerivingAR,jayawardana2017word,Jayawardana2017SemisupervisedIP}. \textit{Legal Ontology Building dataset} and \textit{Legal Ontology Population dataset} was created to experiment with those legal ontologies. However, sentiment analysis in the legal domain can be considered as an area that was not touched until Gamage et al. \cite{gamage2018fast} developed a sentiment annotator specific for the legal domain in 2018. Their approach is based on transfer learning, and it is a domain adaptation task that uses the Stanford Sentiment Annotator\cite{Socher2013RecursiveDM} as the base model. The dataset used for the Stanford Sentiment Annotator consists of 215,154 manually annotated phrases. A study by Xiao1 et al. \cite{Xiao2019CAIL2019SCMAD} introduced the dataset \textit{CAIL2019-SCM} to experiment on similar case matching in the legal domain.

\subsection{Aspect-Based Sentiment Analysis}

A considerable amount of research work has been dedicated to Aspect-Based Sentiment Analysis across several different domains in recent years. There are many publicly available datasets such as movie reviews, products and restaurants to evaluate ABSA tasks. Ganu et al. \cite{ganu2009beyond} created a dataset of restaurant reviews for the task of improving rating predictions. The dataset was annotated on six aspect categories with overall sentiment polarity. The two major steps of Aspect-Based Sentiment Analysis are aspect term extraction and sentiment classification. Hence, this work can not be identified as a fully completed ABSA dataset as it is not using the sentiment polarity of the aspect. It only includes the category of aspect. The semEval 2014 \cite{semeval2014}, an international workshop in the domain of Natural Language Processing (NLP), introduced datasets that are annotated with four fields as aspect term, aspect term polarity, aspect category and aspect category polarity of each sentence. They introduced datasets on restaurant and laptop domains. As a continuation and improvement from semEval 2014, semEval 2015\cite{semeval2015} introduced datasets giving a new definition to the aspect category as a combination of the type of entity and type of attribute for restaurants, hotels and laptop domains. Moreover, semEval 2016\cite{SemEval2016} introduced Multilingual datasets, a total of 39 datasets from 7 domains and 8 languages for the ABSA task. It included datasets of restaurants, hotels, laptops, mobile phones, museums, digital cameras, and telecommunication domains in English, Arabic, Spanish, French, Chinese, Dutch, Turkish, and Russian languages. 

Smadi et al. \cite{smadi2015} created a human-annotated book review dataset in the Arabic language (HAAD). In the annotation process, they used 14 categories along with 4 polarity types as \textit{positive}, \textit{negative}, \textit{neutral}, and \textit{conflict}. For the ABSA of IT product reviews, Tamchyna et al. \cite{Tamchyna2015CzechAS} introduced a dataset in the Czech language. As there were no existing studies on ABSA of Bangla text Rahman et al. \cite{Rahman2018DatasetsFA} created two datasets on cricket and restaurant domains.

To the best of our knowledge, there is no publicly available dataset for the legal domain in the field of ABSA. Consequently, no existing system or research methodologies which intend to effectively analyse the sentiment of legal opinion texts with respect to each legal party in a court case. Therefore, this study aims to address this research gap by introducing the dataset \textit{SigmaLaw-ABSA}  by extracting aspect categories and corresponding polarities of legal texts from the United States supreme court cases. 

\section{Data Collection and Annotation }

\subsection{Data Collection}

This section describes the process of collecting data from legal cases. The court cases were fetched from the \textit{SigmaLaw - Large Legal Text Corpus and Word Embeddings dataset}\footnote{SigmaLaw Dataset -~\url{https://osf.io/qvg8s/}} which contains a large legal data text corpus. The legal data corpus consists of  39,155 legal cases including 22,776 taken from the United States Supreme Court. For the data collection process, about 2000 sentences were gathered to annotate and court cases were selected  without targeting any specific category.
Here we consider 2 types of sentences when preparing the dataset; 

\begin{itemize}[leftmargin=*]
    \item Original Sentence extracted from the court case
    \item Meaningful sub-sentences extracted from the original sentence 
\end{itemize}

Reviews, dialogues, or informal text normally contain shorter sentences where whole sentences can be used in the dataset without any concern. Nevertheless, more complex sources, such as legal documents contain longer sentences and more subordinate clauses~\cite{palau2009}. The subordinate clauses of a sentence can be defined as sub-sentences of that sentence~\cite{ratnayaka2019shift}. 
Sub-sentences can be generated by splitting the sentence using subordinate clauses. In the process of generating sub-sentences, the constituency parser\cite{manning2014stanford} in the Stanford CoreNLP\footnote{Stanford CoreNLP - \url{http://corenlp.run/}} library is used to achieve this task due to the proven performance of the parser \cite{ratnayaka2019shift}.  

\begin{figure*}[!ht]
\centering
\includegraphics[width=0.89\textwidth]{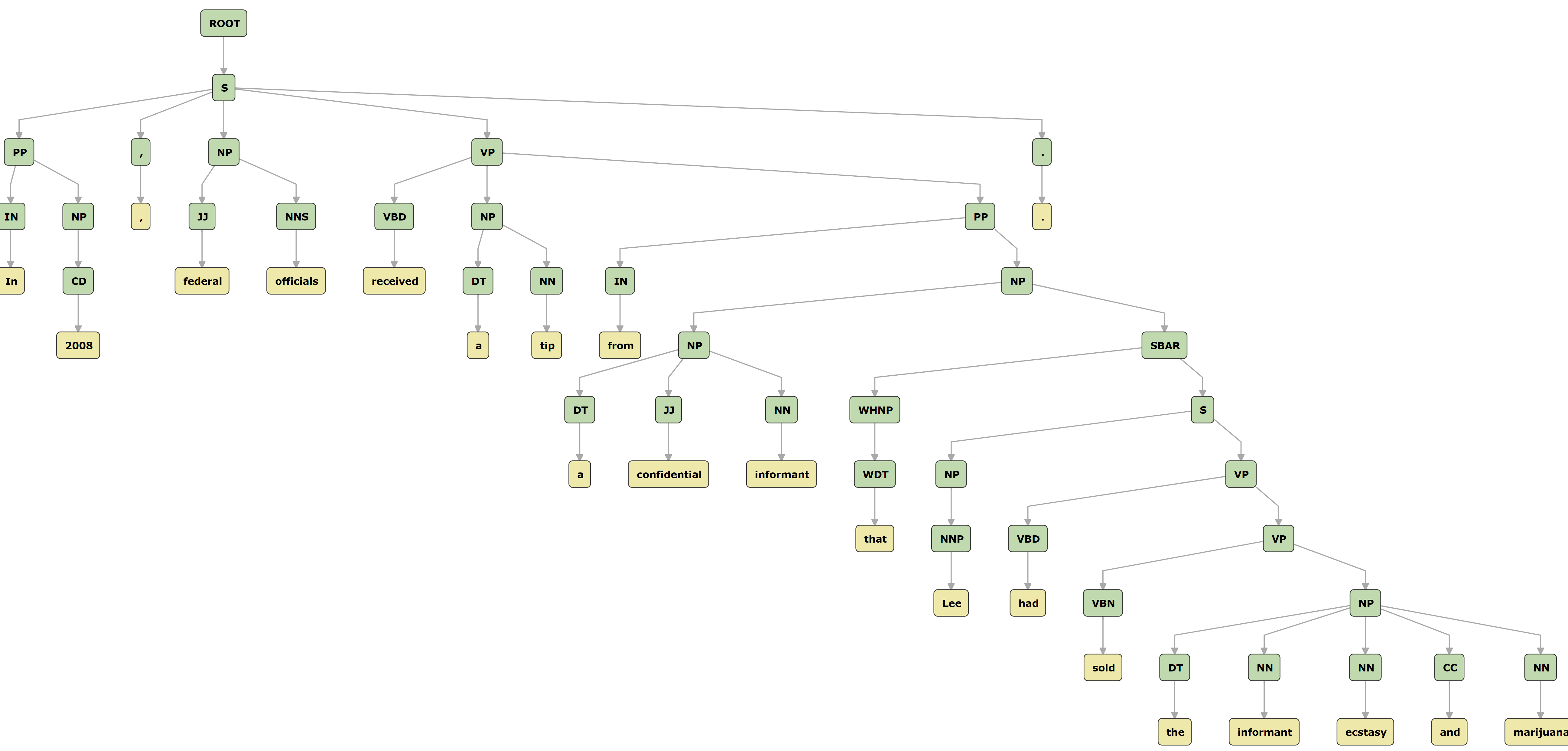}
\caption{Constituency Parser Tree for Example1}
\label{parser}
\end{figure*}

The Constituency Parser returns the parse tree of the sentence after annotation process. The parse tree of the sentence will be built according to the grammatical structure of the sentence. Figure~\ref{parser} illustrates the parse tree for example~\ref{ex1} given above. In the parse tree, subordinating conjunctions label with the SBAR tag. The sub-sentences splitting occurs by recognizing associated terms with the SBAR tag. According to the defined annotation process, 
\textit{"In 2008, federal officials received a tip from a confidential informant"} and \textit{"Lee had sold the informant ecstasy and marijuana"} are the sub-sentences of the Example~\ref{ex1} which splitting by identifying the associated term \textit{that} with the SBAR tag. 

There are some cases which give meaningless sub-sentences after splitting by subordinating conjunctions. In order to extract only meaningful sub-sentences, each sub-sentence is examined by the authors. In this process, all meaningless sub-sentences are eliminated, considering a sub-sentence without a subject as meaningless.

\subsection{Data Annotation}

Reviewing the past literature we found that, there is no publicly accessible tool that assists efficient legal text annotation. The authors, a group of undergraduates and postgraduates in the faculty of law, University of Colombo Sri Lanka were involved in the data annotation process. The annotated dataset contains entities of different parties, their polarities, aspect category (Petitioners and defendants), and the category polarities. The annotators' task was to figure out the aspect categories of each party and give a polarity label for each of them. In this process, we have used polarity types as \textit{Negative}, \textit{Positive}, and \textit{Neutral}. 

Three  human  annotators  analysed the collected  data. As a result, a rating process is needed to determine the final sentiment value. Fleiss' kappa \cite{fleiss1971measuring} was selected to measure for assessing the reliability of agreement between raters (inter-rater reliability). We have obtained the kappa value of 0.59 which belongs to the moderate agreement level and located at the narrow margin of the substantial agreement (0.61). In this dataset, each sentence is annotated under the following fields.

\begin{itemize}[leftmargin=*]

\item \textbf{Party:} Petitioner and defendant are the main parties of a court case and each party may consist of several members. Therefore, in this section, we use list of list data structure and a member who represents the petitioner party or the defendant party goes under the petitioner list (first inner list) or defendant list (second inner list) respectively. Further, when preparing the dataset we considered all required human entity pronouns. 

\item \textbf{Sentiment:} Party based sentiment polarity values are annotated under this field. This field also uses a list of list data structure with the size equal to the list of lists used in the party field. Polarity values for all the parties under petitioner and defendant are listed with respect to listed parties. In the annotation process used polarity types: \textit{Negative}, \textit{Positive}, \textit{Neutral}  as -1, +1, 0, respectively.  

\item \textbf{Overall sentiment:} This section is used to indicate the overall sentiment of the sentences without considering the legal parties related to it. All the sentences are given one of the sentiment levels: \textit{Positive}, \textit{Negative}, \textit{Neutral}  as +1, -1, 0 respectively. 
\end{itemize}

\section{Discussion}

Aspect-Based Sentiment Analysis (ABSA) is a combination of several core sub-tasks. Aspect term extraction, aspect term polarity, category identification, and aspect category polarity can be taken as major sub-tasks. In addition to these sub-tasks, the complexity of the ABSA process depends on the domain and language. As a result of that, much research has been conducted related to these core sub-tasks, domains, and languages. Considering all facts, the datasets of the ABSA field should be multi-tasks. The SigmaLaw-ABSA dataset is prepared to cover all of the above core sub-tasks relevant to the legal domain in the language of English. All sub-tasks are performed on the sentences in court cases. In this section, we discuss the multi-functionality of SigmaLaw-ABSA dataset. 

\subsection{T1 - Aspect Extraction}

For the given legal opinion sentence in a court case, this aspect extraction task concerns the extraction of all persons or organizations that belong to any legal party. Identifying only members of legal parties is a bit of a challenging process. Hence, the aspect extraction model needs better training. As one of its features, this dataset provides all members of parties in sentences.

\subsection{T2 - Aspect Term Polarity}

This task is based on allocating sentiment value (\textit{positive}, \textit{negative}, and \textit{neutral}) to the aspects extracted from the T1 task. The neutral case arises where the content of the sentence is not affected by any party positively or negatively.  

\subsection{T3 - Aspect Category Extraction}

The aspect extraction (T1) task focuses on extracting all persons or organizations belong to any party without any categorization. For lawyers and legal officials, it's beneficial to identify the party of persons or organizations. Therefore, this task deals with categorizing the aspects extracted from a legal opinion text under the petitioner and defendant parties.

\subsection{T4 - Aspect Category Polarity}

This task explores the possibilities of allocating sentiment values for the aspect categories of sentences. T1, T2, and T3 tasks can be applied together for research inspecting how aspects and their polarities influence the polarity of aspect categories. Identifying the sentiment value of the legal texts with respect to the petitioner and defendant parties is an important task when analysing the court cases. The research of identifying sentiment of the parties can be carried out by manipulating the output from the previous tasks. 

\begin{SentenceExample}
\item \begin{Sentence}According to Lee, the lawyer assured him that if deportation was not in the plea agreement, "the government cannot deport you."\end{Sentence}
\label{ex2}
\end{SentenceExample} 

Consider Example 2 taken from Lee v. United States\cite{1977lee} which consists of two legal parties; petitioner Lee and the defendant government. The output from T1, T2, and T3 tasks are \textit{\{Lee, lawyer, government\} , \{positive, neutral, negative\},  and \{petitioner -[Lee, lawyer], defendant-[government]\}}. The output data are sufficient to classify the sentiment values of both petitioner and defendant parties. The output data from tasks T2 and T3 can be combined as follows:  \textit{\{petitioner -[positive, neutral], defendant-[negative]\}}.  We can predict the polarity of parties by adding aspect polarity values of both parties separately. Then we can get a positive value ((+1)+0) for the petitioner party and a negative value (-1) for the defendant party. Researchers can do further experiments on predicting aspect category polarity more accurately. 

After analysing the subtasks of ABSA relevant to the legal domain, we hope that our dataset will be useful to achieve these subtasks. Both T2 and T3 tasks are covered by the field of party which uses the list of lists as its data structure. Here, all aspects belonging to both parties are listed separately. The first and second lists define aspects of petitioner and defendant parties respectively. Utilizing one feature, we have achieved two tasks. Sentiment values of aspects are also defined just like aspects as the list of lists. Task T3 can be performed directly using this dataset. As explained in the task T4, data included in party and sentiment features can be used to carry out those experiments. Besides that, we have indicated the overall sentiment of the sentence as one of the features. The purpose of indicating the overall sentiment is that any researcher can use this dataset to do experiments on identifying the relationship between sentiment analysis and the Aspect-Based Sentiment Analysis fields. Concluding all facts we explored, SigmaLaw-ABSA dataset has proved its multi-functionality.

\section{Dataset Statistics}

This section discloses preliminary results of some statistical analysis performed on this collection of data. After the annotation process, our dataset contains 2000 total sentences which include 1007 full sentences and 993 sub-sentences. Overall, it contains 642 positive 978 negative and 380 neutral sentences. For the Polarity classification of aspect types; petitioner and defendant, it counts 335 positive, 618 negative, 198 neutral for the petitioner and 285 positive, 244 negative, 106 neutral for the defendant. Furthermore, the overall summary of the contents of the dataset is listed in Table~\ref{statistics}. 

\begin{table}[ht]
\setlength{\tabcolsep}{10pt} 
\renewcommand{\arraystretch}{1.3}
\caption{Dataset Statistics}
\label{table:sense2vec}
\centering
\begin{tabular}{|l||c||c||c|}

\hline
\bfseries Polarity                                  & \bfseries Petitioner             & \bfseries Defendant       & \bfseries Total          \\ \hline
Positive                                & 473      &393       &866\\ \hline
Negative                                 &734             &364    &1098  \\ \hline
Neutral                                &254          &143       &397    \\ \hline

\end{tabular}
\label{statistics}
\end{table}

Furthermore, the dataset contains the overall sentiment of sentences as one of its features. The polarity wise details for full sentences and sub-sentences are reported in Table~\ref{overall_statistics}.

\begin{table}[ht]
\setlength{\tabcolsep}{10pt} 
\renewcommand{\arraystretch}{1.3}
\caption{Statics for the Overall Sentiment of Sentences}
\label{table_example}
\centering
\begin{tabular}{|l||c||c||c|}

\hline
\bfseries Polarity                                  & \bfseries Full Sentences             & \bfseries sub-sentences       & \bfseries Total          \\  \hline
Positive                                &357     &285       &642\\ \hline
Negative                                 &517             &461    &978  \\ \hline
Neutral                                &133          &247       &380    \\ \hline
Total                                &1007         &993      &2000   \\ \hline

\end{tabular}
\label{overall_statistics}
\end{table}

\begin{figure}[!ht]
\centering
\includegraphics[width=0.44\textwidth]{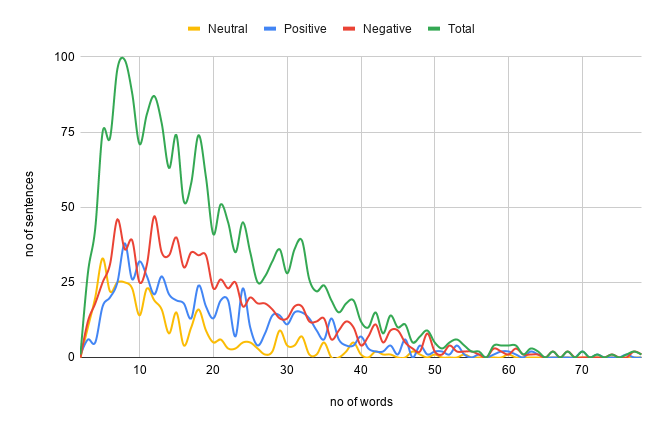}
\caption{Word Count Frequency Plot}
\label{frequency}
\end{figure}

Figure~\ref{frequency} shows word-count frequencies of all sentences along with \textit{positive}, \textit{negative} and \textit{neutral} sentences separately. The sentences of Legal documents are often long and have a complex semantic structure. This plot also highlights that sentences in legal documents are long. Difficulties of handling long sentences and understanding domain-specific phrases emphasize the complexity of the data annotation in the legal domain and the importance of this dataset.

\section{Experiments}

As the dataset contains information related to different tasks described in the discussion section, the dataset can be used to evaluate different models related to the ABSA research area in the legal domain. In this paper, we experimented with the subtask of identification of polarity for each aspect (Task 2 mentioned in the discussion section).  As the first step, preprocessing was performed on the dataset. After this, we classified the sentiment polarity using some popular models available for ABSA using deep learning techniques. These methods include:

\begin{itemize}

\item Target-Dependent Long Short-Term Memory (TD-LSTM)\cite{tang2015effective} - This model extends LSTM (long short-term memory) by considering the target word. It uses two LSTM networks to extract important information from the right and left side of the given target. Then concatenate the hidden states of two LSTM networks and finally classify the polarity labels by feeding them to a softmax layer.

\item Target-Connection Long Short-Term Memory (TC-LSTM)\cite{tang2015effective} -  This model incorporates semantic relatedness of the target with its context words by extending  TD-LSTM with a target connection component.

\item LSTM with Aspect Embedding (AE-LSTM)\cite{wang2016attention} - This model uses an LSTM network and learns an embedding vector for each aspect. Then each word input vector is appended with input aspect embedding.

\item Attention-based LSTM (AT-LSTM)\cite{wang2016attention} - This model uses an  LSTM network with an attention mechanism to incorporate the interaction among words and the target.

\item Attention-based LSTM with Aspect Embedding (ATAE-LSTM)\cite{wang2016attention} - This network incorporates the relationship among words and the target at both the input and attention layer. Each input word vector is appended with aspect embedding and attention mechanism is used along with LSTM.

\item Interactive Attention Networks (IAN)\cite{ma2017interactive} - This model was designed to create representations separately for contexts and targets by interactively learning attention in the target and context. 

\item MemNet\cite{tang2016aspect} - This utilizes a deep memory network and multiple attention is paid to word embedding. For sentiment prediction, the output of the last attention is fed to the softmax layer. Here the results of different attentions are not combined.

\item Content Attention Model for Aspect-Based Sentiment Analysis (Cabasc)\cite{liu2018content} - This model uses two attention enhancing mechanisms. The sentence-level content attention mechanism is responsible for capturing important details of the aspect from a global perspective and context attention mechanism is responsible for taking into consideration at the same time the order of terms and their associations, by combining them into a set of customized memories.

\item Recurrent Attention on Memory (RAM)\cite{chen2017recurrent} - This model first utilizes a bidirectional LSTM to produce memory from the input and weighted the memory slices depending on the relative distance to the aspect. Then multiple attentions on position-weighted memory are non-linearly combined with a recurrent neural network to predict the final sentiment polarity.

\end{itemize}

For word embeddings, we used 300-dimensional pre-trained GloVe Vectors and split the dataset for train/test/validation steps. Table~\ref{performance} summarises the above-mentioned models' performance on the SigmaLaw-ABSA dataset on accuracy and F1 score macro metrics.

\begin{table}[!t]
\renewcommand{\arraystretch}{1.3}
\caption{Performance of models on SigmaLaw-ABSA dataset}
\label{table_example}
\centering
\begin{tabular}{|l||c||c|}
\hline
\bfseries Model & \bfseries Accuracy &  \bfseries F1 score\\
\hline
TD-LSTM & 0.651197
 & 0.564682\\
\hline
TC-LSTM & 0.618271 & 0.543762\\
\hline
AE-LSTM & 0.622754 & 0.558778\\
\hline
AT-LSTM & 0.627245 & 0.559181\\
\hline
ATAE-LSTM & 0.654191 & 0.580193\\
\hline
IAN & 0.633233 & 0.564990\\
\hline
MemNet & 0.538922 & 0.436025\\
\hline
Cabasc &  0.612275 &  0.564300\\
\hline
RAM & 0.663947 & 0.602201\\
\hline
\end{tabular}
\label{performance}
\end{table}

When reviewing the results achieved by each model,
the RAM model's performance on the SigmaLaw-
ABSA dataset is better than the others. The authors of RAM claimed that their approach of using multiple attention mechanisms is better to capture long-distance separated sentiment features and the mechanism is more robust against irrelevant details \cite{chen2017recurrent}. And also the model can handle complications due to the non-linearly combining the results of multiple attentions. Due to the complexity of language and comparatively having long sentences in the legal documents the RAM model outperforms other models on the SigmaLaw-ABSA dataset.

\section{Conclusion}

This study introduces a human-annotated dataset for Aspect-Based Sentiment Analysis for legal opinion texts. To the best of our knowledge, there is no publicly available dataset for the legal domain in the field of ABSA so far made. The  \textit{SigmaLaw-ABSA} dataset consists of 2000 sentences taken from previous court cases. The court cases were collected from the \textit{SigmaLaw - Large Legal Text Corpus and Word Embeddings dataset}. The SigmaLaw-ABSA dataset has been designed to perform various research tasks in the legal domain including aspect extraction, polarity detection, aspect category identification, aspect category polarity detection and this dataset will provide significant importance for the research of the Party-Based Sentiment Analysis for legal opinion texts. Lastly, the dataset is published public and  hosted at OSF\footnote{SigmaLaw ABSA Dataset-\url{https://osf.io/37gkh/}}.

\bibliographystyle{IEEEtran}
\bibliography{paper}

\end{document}